\documentclass[journal]{IEEEtran}
\usepackage{bm}
\usepackage{array}
\usepackage{color}
\usepackage{amsmath}
\usepackage{algorithmic}
\usepackage{array}
\usepackage{algorithm}
\usepackage{epsf}
\usepackage{amsthm}
\usepackage{graphicx}
\usepackage{amssymb}
\usepackage{wrapfig}
\usepackage{mathtools}
\usepackage{subfigure}
\usepackage{multirow}
\newcommand{\reffig}[1]{Fig.~\ref{#1}}

\usepackage{cite}

\ifCLASSINFOpdf
\else
\fi
\begin{document}
%

\title{Compare Contact Model-based Control and Contact Model-free Learning: A Survey of Robotic Peg-in-hole Assembly Strategies} 
%
%
%

\author{Jing~Xu,~\IEEEmembership{Member,~IEEE,}~
        Zhimin~Hou,~
        Zhi~Liu~
        and~Hong~Qiao,~\IEEEmembership{Fellow,~IEEE}
\thanks{
Jing Xu, Zhimin Hou and Zhi Liu are with the State Key Laboratory of Tribology, Beijing Key Laboratory of Precision/Ultra-Precision Manufacturing Equipment Control, Department of Mechanical Engineering, Tsinghua University, Beijing 100084, China.
}
\thanks{H. Qiao is with the State Key Laboratory of Management and
Control for Complex Systems, Institute of Automation, Chinese Academy of
Sciences, Beijing 100190, China.}
\thanks{Correspondence e-mail: jingxu@tsinghua.edu.cn,  hong.qiao@ia.ac.cn}
}

%
%

\markboth{March~2019}%
{Shell \MakeLowercase{\textit{et al.}}: Bare Demo of IEEEtran.cls for IEEE Journals}
%

\maketitle

\begin{abstract}
In this paper, we present an overview of robotic peg-in-hole assembly and analyze two main strategies: contact model-based and contact model-free strategies. 
More specifically, we first introduce the contact model control approaches, including  contact state recognition and compliant control two steps. 
Additionally, we focus on a comprehensive analysis of the whole robotic assembly system. 
Second, without the contact state recognition process, we decompose the contact model-free learning algorithms into two main subfields: learning from demonstrations and learning from environments (mainly based on reinforcement learning). 
For each subfield, we survey the landmark studies and ongoing research to compare the different categories. 
We hope to strengthen the relation between these two research communities by revealing the underlying links. 
Ultimately, the remaining challenges and open questions in the field of robotic peg-in-hole assembly community are discussed. The promising directions and potential future work are also considered. 
\end{abstract}


%
\IEEEpeerreviewmaketitle

\section{Introduction}
%
%
%
%

\IEEEPARstart{R}{obotic} assembly as the essential components of industrial applications has been studied for a long time. In this work, we look at the most common problem of robotic assembly: peg-in-hole assembly, which is the basis of a wide range of component assemblies\cite{kuangenjamming}\cite{su2017sensor-less}. 
Robotic peg-in-hole assembly has been extensively researched and applied in various fields from large-scale object assembly, such as aviation components\cite{wan2017optimal}\cite{qiao2016largescale}, engines\cite{su2012new} and windshields assembly to small-scale components, such as mold casting manufacturing\cite{visionformultiplepeghole}, electronic components\cite{su2012sensor} and even microproduct\cite{chang2011visual} assembly. 

\subsection{The development of robotic peg-in-hole assembly}

\begin{figure}[!t]
\centering
\includegraphics[width=3.5in]{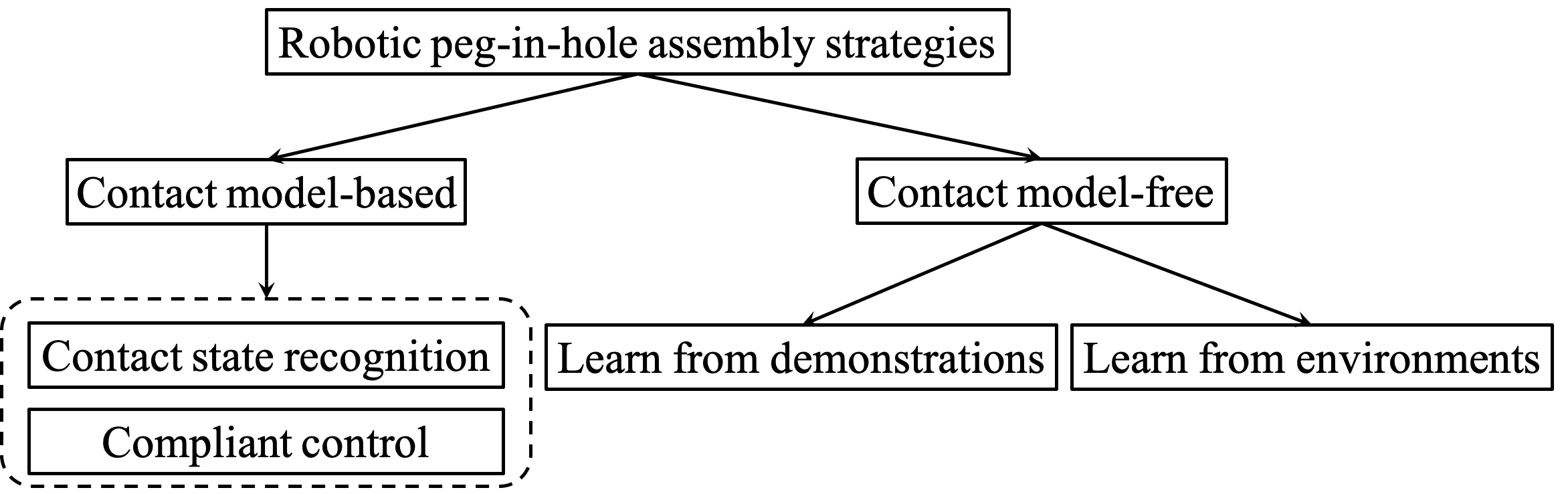}%
\caption{Systematic of robotic peg-in-hole assembly strategies.} 
\label{fig_architecture_this_paper}
\end{figure}

Many academic and industrial researchers have focused on promoting the robotic peg-in-hole assembly on the basis of classical conditioning learning with conventional compliant control strategies\cite{lefebvre2005active}, observational learning with learning from demonstrations\cite{zhu2018robot}, and operant conditioning learning with learning from environments\cite{sutton2018reinforcement}. 
In this work, we decompose the existing peg-in-hole assembly strategies into contact model-based and contact model-free two categories, as illustrated in \reffig{fig_architecture_this_paper}. Furthermore, the contact model-free strategies can be further subdivided into learning from demonstrations and learning from environments. 

\subsubsection{Conventional contact model-based strategies}
The concept of conventional contact model-based strategies relies on the contact model analysis and decomposes the peg-in-hole assembly into two steps: contact-state recognition and compliant control. 
The compliant control strategies are preprogrammed by humans according to contact state recognition by analyzing the underlying friction and contact model\cite{lefebvre2005active}. 
Researchers have made efforts to use contact model based strategies to solve a wide range of peg-in-hole assembly problems with special requirements and high complexity in autonomous industrial manufacturing. 
For instance, the best methods for performing high-precision assembly\cite{su2017sensor-less}\cite{rlhighpercision}\cite{2018modelfreelearning} and large-scale component assembly\cite{zhi2014largescale}\cite{qiao2016largescale}\cite{wan2017optimal} with the limited sensors have been widely investigated. 
Additionally, methods for deriving an assembly control strategy that can cope with complicated multiple peg-in-hole assembly flexibly has attracted considerable attention from researchers\cite{kuangenjamming}\cite{zhiminfeedback}\cite{hou2018learning}. 

To date, most of the published research worked on the peg-in-hole assembly has focused only on optimizing the separate stages. 
On the one hand, contact-state recognition has been explored to improve the success rate of recognition through theoretical analysis\cite{whitney1982jamming} and statistical techniques\cite{jasim2017contact} without caring about assembly implementation. 
On the other hand, to enhance the efficiency and stability with little contact forces during assembly, some optimization approaches have been applied to improve the performance of compliant control strategies directly\cite{nn2002analysis}\cite{tang2016autonomous}\cite{hou2018learning} without considering the assistance of contact state recognition results. 
However, less research has focused on analyzing the relations between the contact state recognition and control strategies and on integrating these two stages \cite{lefebvre2005active}.

\subsubsection{Learning from demonstrations~(LFD)}

As autonomous robotic peg-in-hole assembly techniques progress, the compliant assembly control strategies are expected to perform more complicated assembly with higher degrees of compliance in a unstructured and nonstationary environments. 
It is possible for the preprogramming contact model based compliant control strategies to take into account all the possible assembly situations in advance. 
From the perspective that human beings are capable of handing various complicated assembly flexibly and unpredictably, LFD methods\cite{argall2009surveyforlearingdemonstration} are based on the idea that the assembly behaviors can be learned by interpreting humans demonstrations without preprogramming and have been developed to solve peg-in-hole assembly in the recent years\cite{wan2017optimal}\cite{tang2015learning}\cite{tang2016teach}. 
To imitate the compliant behaviors of human beings for peg-in-hole assembly, in addition to the assembly motion path, the force control strategies are also taken into account\cite{tang2015learning}\cite{tang2016teach}. 
A comprehensive survey of LFD methods was presented in \cite{argall2009surveyforlearingdemonstration}, which contributed a structure with demonstration \emph{gathering} and policy \emph{deriving} two phases. 
Zhu and Hu\cite{zhu2018robot} surveyed the LFD techniques applied in general robotic assembly and introduced the whole demonstrations assembly system. 
Kyrarini et al.\cite{kyrarini2018robot} examined and compared several modeling methods of human demonstrations for industrial assembly tasks. 

The LFD method is an effective learning algorithm used to solve robotic assembly problems. However, few studies have analyzed the challenges of LFD methods for peg-in-hole assembly scenarios. 
Furthermore, less research has focused on improving the ability of adaptation to environmental changes, uncertainties and generalization in new assembly situations. 

\subsubsection{Learning from environments~(LFE)}

In contrast to improving the generalization of LFD methods, current robots are expected to recognize the surrounding environments actively and to learn the assembly skills incrementally, similar as human beings. 
\emph{Reinforcement learning}~(RL) based methods hold great promise for achieving such performance, and these methods enable agents to learn behaviors through integration with the surrounding environments and ideally by generalizing to unseen scenarios or tasks\cite{sutton2018reinforcement}. 
To solve the inherent difficulties in behavior modeling and the generalization, an adaptable and robust control system was developed not only through learning from expert demonstrations but also incremental learning. 
With the development of the \emph{artificial intelligence} techniques, especially \emph{deep learning}, 
typical RL based learning approaches, especially model-free learning algorithms, have been extensively applied to perform complex manipulations\cite{jan2013reinforcement}\cite{levine2015learning}, including robotic peg-in-hole assembly\cite{rlhighpercision}\cite{rl2018peginhole}. 

It is widely accepted that it is possible to apply model-free RL algorithms in real-world robotic assembly tasks at the expense of data efficiency. 
To enhance the practicalities, many studies focusing on incorporating typical model free RL approaches with the prior knowledge or expert demonstrations have been published\cite{zhiminfeedback}\cite{2018learningfromCAD}\cite{2018modelfreelearning}. 
Recently, there has been increasing interest in the development of model-based RL in the robotics community. 
For example, transition dynamics models have been utilized to derive the feedback rewards or optimal actions and have been investigated in \cite{levine2015learning}\cite{polydoros2017survey}\cite{2018modelfreelearning}. 
However, for robotic peg-in-hole assembly, 
it is not clear how to fuse the existing knowledge into a model-free learning process naturally. 
Although, some papers have worked on the comparison and combination of model-based and model-free RL learning algorithms for robotic applications\cite{polydoros2017survey}, no survey has yet explored the relation between model-based RL learning algorithms with the conventional theoretical contact model and the implicit model learning from demonstrations. 

\subsection{The motivation and purpose of this paper}

Although numerous studies on robotic assembly have been published, there is still no paper surveying the existing research, including both the contact model-based and two kinds of contact model-free assembly strategies for peg-in-hole assembly. 
To the best of our knowledge, few studies have compared conventional contact model-based strategies and contact model-free algorithms. 
Therefore, one motivation of this paper is to survey the existing assembly strategies and group them as shown in \reffig{fig_architecture_this_paper} for the first time. 
Another motivation is to exploit the underlying relations between different assembly strategies and to explore the promising solutions by combining the strengths of contact model-based and contact model-free control strategies. 
Consequently, to make existing peg-in-hole research tractable, we attempt to give a fairly complete overview with the following goals
\begin{enumerate}
    \item This paper surveys the state-of-the-art research and ongoing developments of robotic peg-in-hole assembly and identifies promising approaches. 
    \item This paper provides a novel and clear grouping method to analyze the existing research completely with comprehensive insights. 
    \item This paper explores the underlying relationship between traditional contact model-based control strategies and contact model-free learning algorithms and proposes the promising solutions. 
    \item This paper highlights the remaining challenges of the existing approaches and identifies  open questions for future research. 
\end{enumerate}

The remainder of the paper is organized as follows: Section~\ref{S:2} introduces the whole robotic peg-in-hole assembly system. 
Section~\ref{S:3} analyzes the contact model-based control strategies in detail. 
Section~\ref{S:4} surveys and compare two contact model-free learning algorithms. 
Section~\ref{S:5} concludes with a discussion of the open questions and potential future research directions.


\section{Robotic peg-in-hole assembly system}\label{S:2}
In this section, we briefly introduced the construction of a robotic peg-in-hole control system briefly, as shown in \reffig{fig_framework_assembly_system}, which consists of three components: mating parts, the sensing system and manipulators. 
Generally, the holes are fixed and the manipulators grab the pegs to complete the parts mating according to the feedback from the sensing system. 

\subsection{Mating parts}
\begin{figure}[!t]
\centering
\subfigure[]{\includegraphics[width=3.1in]{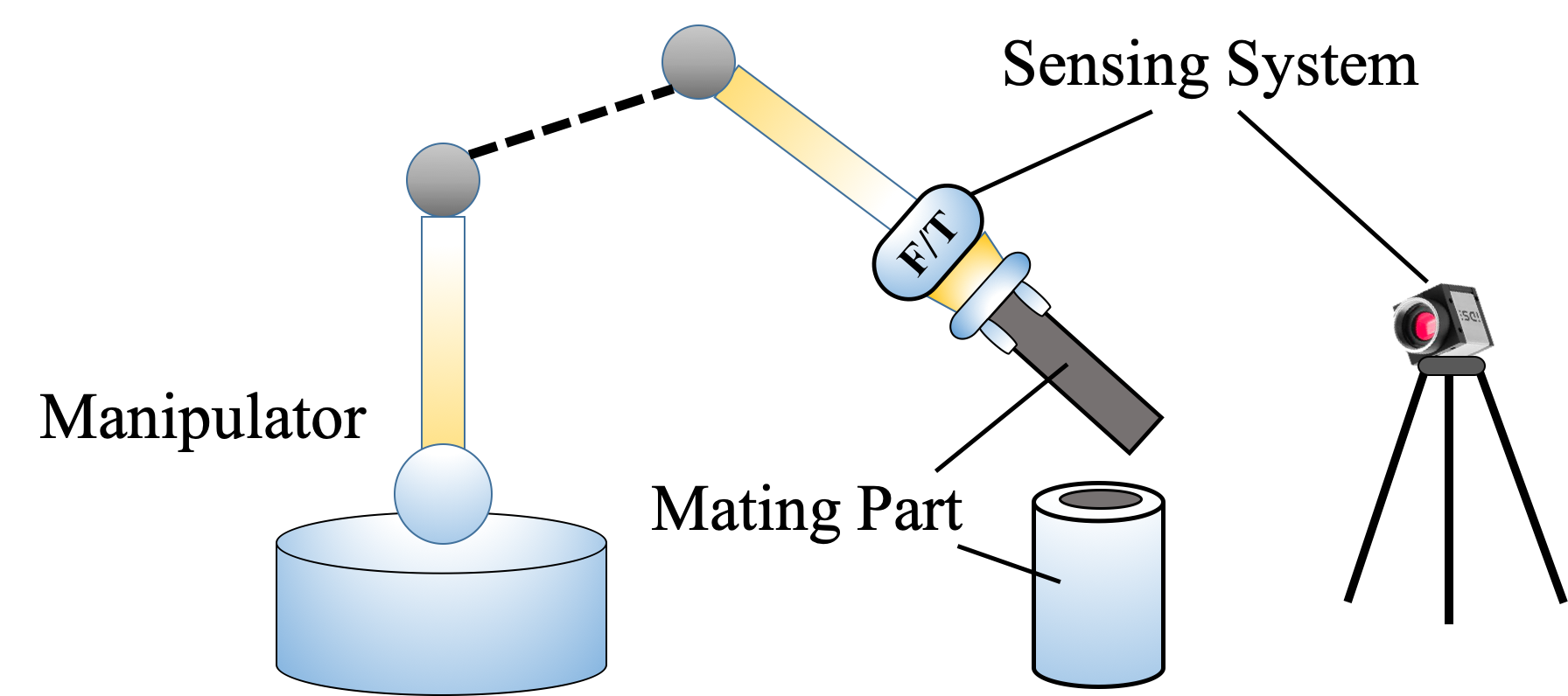}%
\label{fig_framework_assembly_system}}
\hfill
\subfigure[]{\includegraphics[width=3.3in]{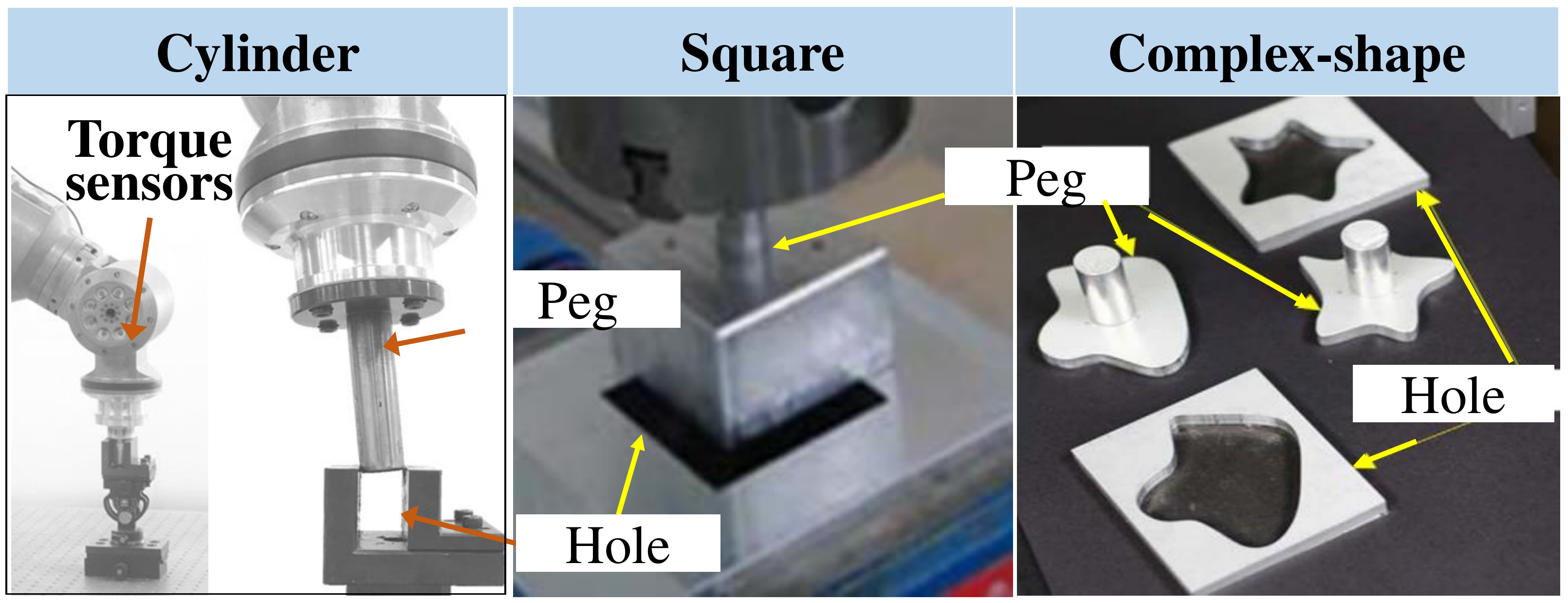}%
\label{fig_shape_case}}
\hfil
\subfigure[]{\includegraphics[width=3.3in]{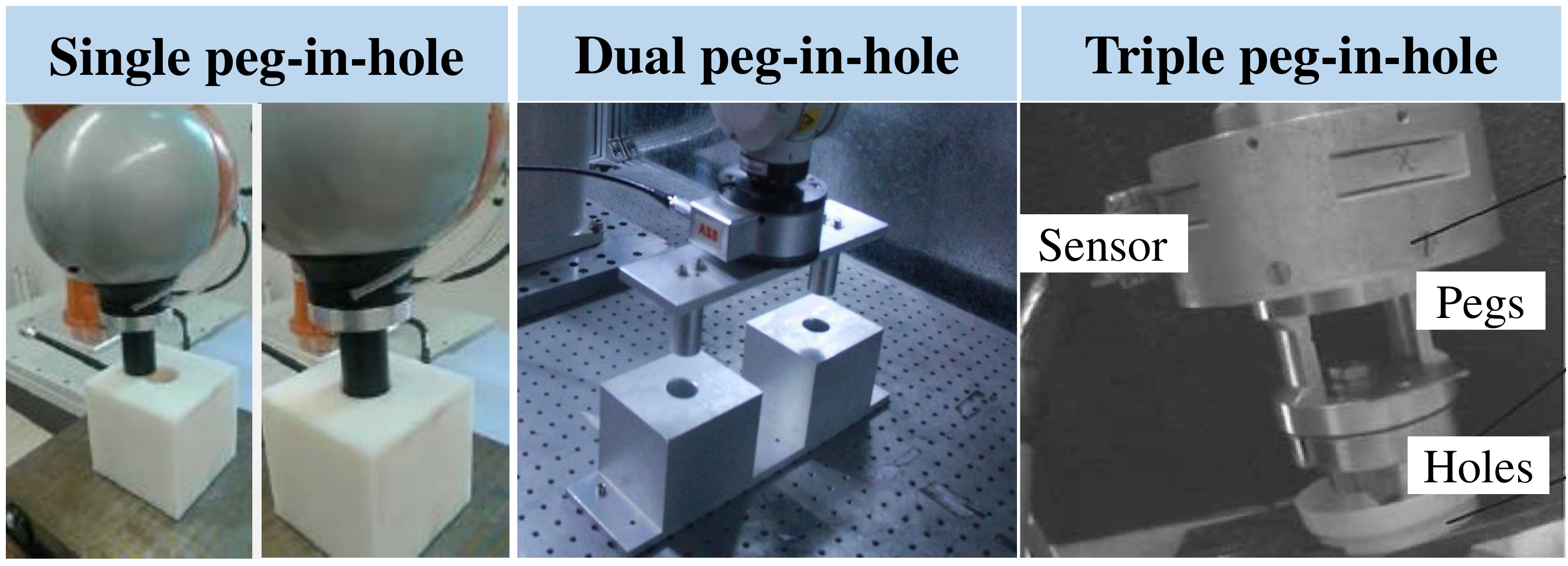}%
\label{fig_pegs_number_case}}
\caption{Introduction of a robotic peg-in-hole assembly system. (a)Setup of the robotic peg-in-hole assembly system. (b)~Figures from Ren et al.\cite{ren2018learning}, Kim et al.\cite{kim2014holedectect} and Song et al.\cite{2014forceguide}. (c)~Figures from Jasim et al.\cite{jasim2014position}, Xu et al.\cite{zhiminfeedback} and Fei and Zhao  \cite{fei2003assembly}}. 
\label{fig_peg_hole_assembly_examples}
\end{figure}

The mating parts, as shown in \reffig{fig_framework_assembly_system}, are the assembly components and include the pegs and holes. 
According to the geometrical features shown in \reffig{fig_shape_case}, a cylindrical peg-in-hole system is the basic assembly problem and has been extensively studied. 
Complex-shape peg assembly is also used in some special cases, including square pegs\cite{park2013intuitive}\cite{kim2014holedectect}, pegs with key slots and pegs with complex shapes\cite{2014forceguide}.  
In addition, according to the number of peg-hole mating pairs, the research work can be decomposed into single peg-in-hole\cite{wan2017optimal} and multiple peg-in-hole assembly\cite{fei2003assembly}\cite{kuangenjamming} scenarios, as shown in \reffig{fig_pegs_number_case}. 
The complexity of assembly increases as the contact states of multiple peg-in-hole scenarios become more complicated. 

\begin{table*}[!t]
\centering
\caption{Conclusion for different sensing system.}
\begin{tabular}{l l l l}
\hline
\textbf{Category} & \textbf{Types} & \textbf{Characteristics} & \textbf{Methods}\\
\hline
Vision & $\bullet$ Camera(2D, stereo), laser tracker & $\bullet$ Contact-less, low-resolution & $\bullet$ Boundary detection, visual servo strategy\\
Force & $\bullet$ F/T, torque & $\bullet$ Monitors contact force & $\bullet$ Blind search strategy, impedance control, force-based position control\\
Sensor-less & $\bullet$ Joint current, joint encoder & $\bullet$ Low cost, no installation & $\bullet$ ARIE-based inserting, human-like exploration searching\\
\hline
\end{tabular}
\label{table_sensing_system}
\end{table*}

The scale of assembly components corresponds to the application and ranges from macroassembly for large aviation parts to microassembly for electronic components in circuit board. 
The clearance between peg and hole also differs with the requirements of the assembly scenarios. 
In some high-precision scenarios\cite{rlhighpercision}\cite{tang2015learning}\cite{kuangenjamming}, the clearance may be below the solution and accuracy of the robot, which is typically in the range of 0.02 up to 0.2~mm. 
In addition to the rigid pegs with high stiffness values\cite{zhang2017force}, some flexible peg-in-hole components composed of plastics\cite{kuangenjamming} and wood are also used. 
The clearance and hardness of the mating surfaces change the nature of the part mating tasks\cite{2012contactsvm}, including various degrees of complexity. 

\subsection{Sensing system}
In the case of robotic peg-in-hole assembly, the sensing system is used to acquire feedback from environment, similar to human sight and tactile sensing. 
Sensing systems based on two types of sensors or other feedback are surveyed, and the corresponding characteristics are as follows. 

\subsubsection{Vision sensors}
 
2D cameras are widely used for coarse localization by extracting the boundaries of holes from images\cite{miura1998vision}. 
Maker points captured by 2D cameras were used to calculate the pose~(position and orientation) of pegs in\cite{wan2017optimal}. 
Image-based visual servo systems were designed to track the accurate hole position based on the extracted features\cite{pauli2001vision} \cite{wang2008microassembly}. 
For high-speed microscale peg-in-hole assembly, Chang et al.\cite{chang2011visual} and Huang et al.\cite{huang2013visualservoing} proposed position-based visual servo systems with fast convergence guarantees based on the image calibration method and limited calibration. 
In contrast to images, stereo cameras~(Kinect) have been applied to capture 3D point data to estimate the accurate 3D poses of mating parts\cite{abu2014solving}\cite{park2017compliance}. 
Additionally, laser trackers, such as the high-precision and contact-less tools have been employed to enhance the position accuracy of large-scale peg-in-hole assembly systems\cite{zhi2014largescale}\cite{qiao2016largescale}. 

\subsubsection{Force sensors}

Position controllers based on vision sensors might produce large contact forces due to the position errors. 
Therefore, force feedback can be utilized not only to monitor the assembly process, but also to accommodate the position uncertainty. 
The force feedback referred to as wrench signals~(forces and moments) can be acquired from an external force-torque~(F/T) sensor equipped on the end-effector of the robot\cite{zhiminfeedback} and from torque sensors integrated into the robot joints\cite{lee2014active}\cite{ren2018learning}, as shown in \reffig{fig_peg_hole_assembly_examples}. 

In general, force feedback is used for compliant control strategies in passive and active ways. 
As a passive example, auxiliary mechanical devices such as remote-center-compliance~(RCC)\cite{whitney1982jamming}\cite{sturges1996RCC} and a variant \cite{xu2015robust} (composed of springs and dampers) attached to the end-effector were applied to accommodate the contact forces. 
Active compliant control strategies aim to control the assembly motions of robots actively based on force feedback \cite{su2017sensor-less} and have been widely surveyed for more than 30 years. 
Wrench signals from sensors have been applied to recognize the contact-state and generate the low-level commands for position control of robots\cite{lefebvre2005active} through impedance control\cite{hogan1984impedance} and force-based position control\cite{raibert1981hybrid}. 

\subsubsection{Sensorless systems}

To eliminate the limitations of sensor frequency, sensor installation and measurement error, De Luca and Mattone\cite{de2005sensorless} and Lee and Park\cite{lee2014active} worked on an efficient sensor-less active compliant control systems without external F/T sensors, in which the wrench signals were approximated according to the current of joint motors. 
Additionally, the poses of pegs attached on the end-effector were interpreted by the encoder in the robots. 
Based on the pose information, robotic peg-in-hole assembly strategies driven by the environmental constraints, such as attractive regions in the environment~(ARIE) without force-sensor feedback have been investigated recently\cite{qiao2015concept}\cite{qiao2017iros}. 

Consequently, both vision sensors and force sensors have the strengths and shortcomings, as summarized in Table.~\ref{table_sensing_system}. 
Most vision sensors are appropriate for peg-in-hole assemblies with larger clearances and weak contact forces\cite{huang2013visualservoing}. 
Nevertheless, force-based assembly control strategies have been explored in high-precision assembly systems with small clearances\cite{rlhighpercision}\cite{2018modelfreelearning}, large-scale components assembly systems with large contact forces\cite{hou2018learning}\cite{kuangenjamming} and complex-shaped parts assembly systems with complex contact forces\cite{song2016guidance}\cite{dietrich2010contact}. 
To combine the strengths of vision and force sensors, the hybrid sensing systems have also been developed in \cite{xie2009hybridsensor}\cite{2004datafusion}. 


\subsection{Manipulators}

The manipulators in peg-in-hole assembly can be industrial robots~(such as those produced by the companies ABB and KUKA) with 6 degrees-of-freedoms~(DOFs)\cite{kuangenjamming}\cite{zhang2017force}, which are used to perform assembly requiring large forces and moments. 
Furthermore, in recent years, high-compliance robot manipulators~(Baxter, PR2) with 7 or 8 DOFs have been developed and can perform manipulation problems more flexibly and safely\cite{park2017compliance}. 
Generally, most robotic peg-in-hole assembly environments involve fixed holes, as shown in \reffig{fig_peg_hole_assembly_examples}, and manipulators are used to grab the pegs to control the motions~(translation and rotation) in Cartesian space or joint angles in joint space. 

The robotic peg-in-hole assembly for inserting the pegs to the desired depth of holes generally consists of two main phases: \emph{searching} and \emph{inserting}. 
For the \emph{searching} phase, the localization of holes should be identified, which is an essential step for the following \emph{inserting} phase in real-world scenarios. 
Image-based boundary extraction techniques\cite{chang2011visual}, visual servo tracking approaches\cite{wang2008microassembly} and blind search strategies based on the designed human-like searching path\cite{chhatpar2001search} and the force feedback\cite{kim2014holedectect} have been applied to locate holes and track the position. 
For the \emph{inserting} phase, the assembly actions involve not only motion but also the applied external forces. 
In contrast to the searching phase, the \emph{inserting} phase is more complicated; most researchers only focus on active control strategies for the \emph{inserting} phase and neglect the \emph{searching} phase\cite{tang2016autonomous}\cite{zhang2017force}\cite{hou2018learning}. 
Therefore, in this work, all of the following analysis of robotic peg-in-hole assembly are for the \emph{inserting} phase. 


\section{Contact model-based control strategies}\label{S:3}

\begin{figure}[!t]
\centering
\subfigure[]{\includegraphics[width=2.9in]{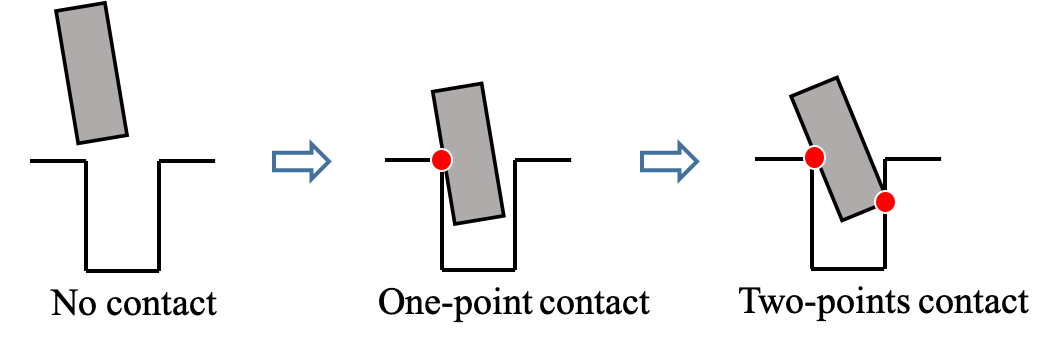}%
\label{fig_contact_state}}
\hfil
\subfigure[]{\includegraphics[width=2.9in]{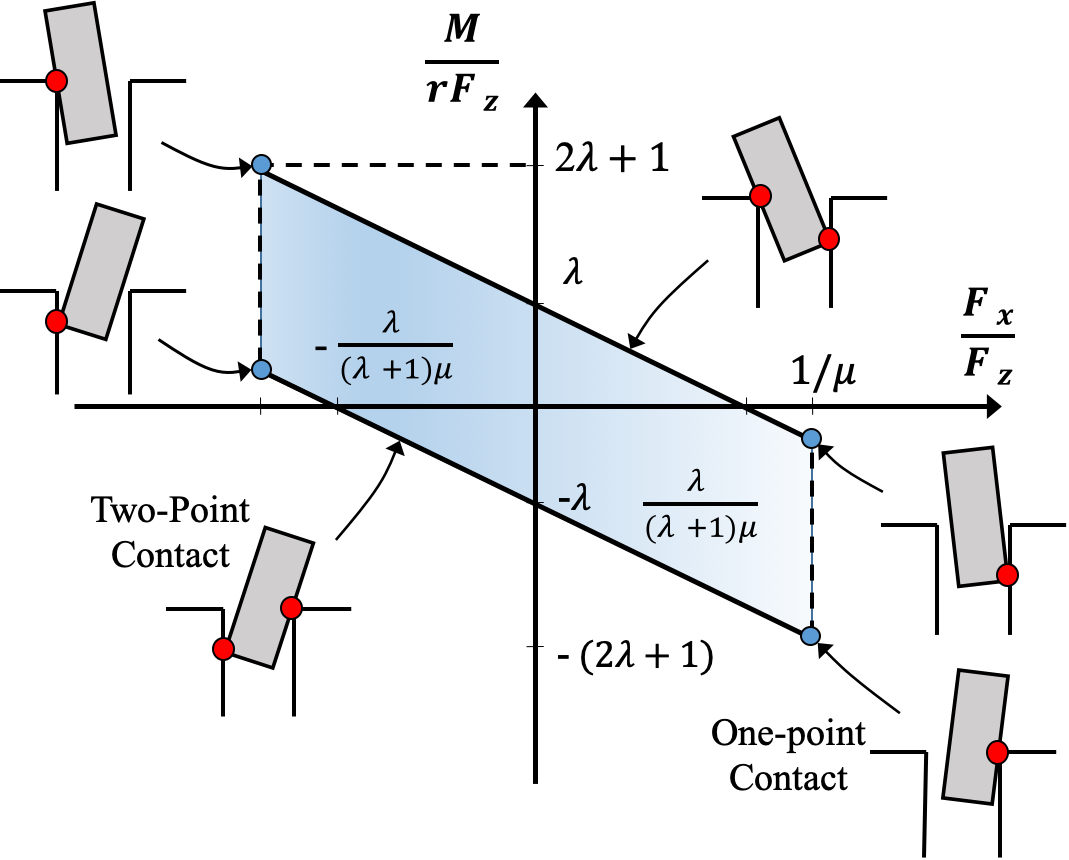}%
\label{fig_jamming_diagram}}
\caption{Contact state analysis and jamming diagram for single peg-in-hole assembly~(Whitney~\cite{whitney1982jamming}).}
\end{figure}

\begin{figure*}[!t]
\centering
\includegraphics[width=7.0in]{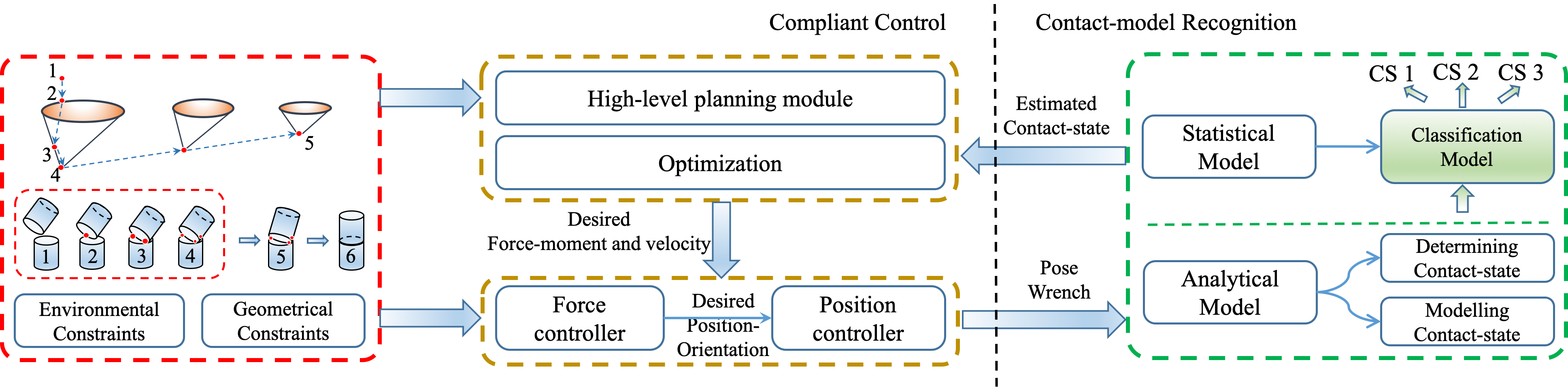}
\caption{Framework of contact model-based strategies.}
\label{fig_control_framework}
\end{figure*}

The assembly process is a constrained motion with geometrical and environmental constraints. 
The contact constraints between mating parts can be represented as topological contact states\cite{291962}\cite{xiao1998contact}\cite{10.1007/978-3-642-83625-1_17}. Thus, the overall assembly process can be described as a sequence of transitions between the contact states. 
For instance, as shown in \reffig{fig_contact_state}, a single peg-in-hole insertion process is formulated as the transitions between no contact, one-point contact and two-point contact. 
The contact model-based strategies for robotic assembly shown in \reffig{fig_control_framework} generally include two steps: contact sate recognition and compliant control. 

The general idea of contact state recognition is to determine the contact constraints according to the observations, such as wrench signals and pose information. 
In this work, we analyze and decompose the existing research for contact state recognition into two categories: the analytical model\cite{whitney1982jamming} and the statistical model\cite{2012contactsvm}. 
The analytical model relies on the analysis of the geometrical and environmental constraints. 
The statistical model has been extensively developed in recent years and estimates the contact state through learning the pattern from the collected samples directly without the need for information about the tasks.

\begin{table*}[!t]
\centering
\caption{Comparison for contact states recognition methods with statistical model.}
\begin{tabular}{l l l l l}
\hline
\textbf{Categories} & \textbf{Success rate(\%)} & \textbf{Computational time($s$)} & \textbf{Advantages} & \textbf{Disadvantages}\\
\hline
GMM, ~DSM-GMM & 94.4 & 18.795 & $\bullet$ Fit distribution of samples & $\bullet$ Sensitive to the initial setting\\
SVM, ~SVM-FIM & 64.2 & 70.719 & $\bullet$ Excellent generalization & $\bullet$ Sensitive to missing sample\\
CFC, ~GS-FCA\ & 27.3/65.9 & 0.002/237.307 & $\bullet$ Fuse prior knowledge & $\bullet$ Only solve simple case\\
SGB\ & 60.7 & 92.083 & $\bullet$ Without defining parameters & $\bullet$ Sensitive to unseen samples\\
HMM & \-- & \-- & $\bullet$ Eliminating time-varying uncertainties & $\bullet$ Sensitive to gain value\\
NNs & \-- & \-- & $\bullet$ Easy-implementation & $\bullet$ Less data-efficient\\
\hline
\end{tabular}
\label{table_staistical_model}
\end{table*}

\subsection{Contact state recognition with analytical model}

The contact state recognition with analytical model\cite{xiao1998contact} generally includes two steps: contact state modeling and contact state determining. 

\subsubsection{Contact state modeling}

Contact states are modeled according to the mating features, and the most commonly used features are the force constraints between mating parts. 
Whitney \cite{whitney1982jamming} proposed a quasi-static model to clarify the relation between force and geometrical constraints for a single peg-in-hole assembly.  
Additionally, two possible ill contact situations~(\emph{wedging} and \emph{jamming}) are analyzed. Jamming, which often leads to insertion failure, represents the conditions in which the applied forces/moments of the peg are in the wrong proportions. 
As shown in \reffig{fig_jamming_diagram}, a jamming diagram is drawn to analyze the jamming conditions of the overall assembly process\cite{whitney1982jamming}. 
Based on this idea, 
Sathirakul and Sturges\cite{sathirakul1998jamming} and Fei and Zhao\cite{fei2003assembly} enumerated contact states and presented a three-dimensional analysis of jamming conditions for multiple peg-in-hole assembly. 

In contrast to the quasi-static model and rigid body assumption, some researchers have focused on dynamical or flexible models, which are closer to real assembly situations. 
Hsin-Te et al.\cite{liao1998analysis} derived a general form of impact equations for an industrial manipulator performing peg-in-hole assembly using Lagrange's impact model. 
Xia et al.\cite{xia2006dynamicanalysis} established a three-dimensional jamming analysis based on a compliant elastic contact dynamic model and designed several no-jamming and no-wedging assembly strategies by analyzing the free contact conditions. 

\subsubsection{Contact state determination}

The contact states are determined by calculating the similarity between the observed actual and modeled contact states. 
As the contact state recognition moves forward, the contact states are modeled with uncertain parameters to accommodate the error in the contact model and to enhance the robustness to environmental uncertainties. 
\emph{Kalman filters}~(KF)\cite{lefebvre2005online} and \emph{particle filters}\cite{gadeyne2005bayesian} have been utilized to estimate the geometrical parameters for better recognition of the contact state and state transitions. 
In addition, instead of determining the contact state by calculating the similarity, classification algorithms, such as \emph{support vector machine}~(SVM), have been applied to determine the contact state \cite{2012contactsvm}. 

In conclusion,  
the analytical model is sensitive to uncertainties~(such as that in the position of parts, the rigidity or elasticity of the assembly system and the friction model), and no perfect model can be adapted automatically to different environments. 
The aforementioned methods with analytical models relying on force constraint analysis will become more complicated in assembly systems with uncertain mating features and changing jamming conditions. 
Consequently, only a partial model can be achieved, and the generalization to new assembly scenarios is limited. 
Another drawback is that the variables of an analytical model can only be determined based on the observed contact states and past transitions. 

\subsection{Contact state recognition with statistical model}

In contrast to recognition based on an analytical model, 
contact state recognition with a statistical model without considering the possible uncertainties is formulated as a classification problem given the possible contact states. 
The contact state can be classified through the advanced statistical techniques such as \emph{fuzzy classifiers}~(FC),~\emph{neural networks}~(NNs),~SVM,~\emph{Gaussian mixtures model}~(GMM) and~\emph{hidden Markov models}~(HMMs). 

Conventional fuzzy classifiers~(CFC) have been applied in contact state recognition by accommodating the uncertainties based on prior knowledge without the geometrical information on the pegs\cite{xiao1998contact}\cite{1998conceptfuzzy}\cite{2000fuzzyanalysis}. 
In these scenarios, the output contact state is decided through the following fuzzy \emph{if-then} rules
\begin{equation}
    if ~ X_1^k ~ is ~ A_i^1, ... and ~ X_j^k is ~ A_i^j, then ~ Y = S_i
\end{equation}
where $X_j^k$ denotes the $j$th component of the $k$th input observation signal, $A_i^j$ is the antecedent membership function of the $j$th input component for the $i$th contact state $S_i$.   
To enhance the robustness of the fuzzy system, the \emph{gravitational search}~(GS) algorithm is employed to tune the fuzzy rules of each model\cite{2013lmsfuzzy}. 
CFC is able to solve the simple classification problem through the designed fuzzy logic controller with little computing time. 

NNs have been developed for a long time and were used to map the nonlinear relationship between the input force information and output contact states\cite{nn2002analysis}. 
Compared to FC-based methods, the implementation of NNs is feasible without handcrafted extraction features and fuzzy rules. 
NNs have shown competitive classification performance in recent years with sufficient computing resources and samples. 
The main issue is that the trained classification model cannot be generalized to the scenarios with different dimensions of inputs due to the fixed network architecture. 
The performance compared to CFC was analyzed in \cite{nn2002analysis}, and both of them have advantages and disadvantages. 
Additionally, numerous studies focusing on integrating the flexibility of fuzzy set theory and the approximation ability of NNs have been performed\cite{son2002optimal}\cite{son2001neuralfuzzy}. 
For training the classifiers with NNs, the input observed information, including measured wrench signals and pose data, usually requires preprocessing, such as normalization or uniform discretization. 

SVM techniques through reducing the actual risk and confidence interval for correct classification, have been demonstrated to be suitable and applicable for real-world recognition with generalization to unknown environments\cite{2012contactsvm}. 
Previous work has proposed a practical contact state recognition framework, in which the input observations are processed through discrete wavelet transform~(DWT) and the contact states are acquired through existing analytical models. 
A fuzzy inference mechanism~(FIM) with an adaptive classifier boundary generated by SVM was used to classify the contact states of the peg-in-hole assembly sequence\cite{2014fuzzysvm}. 

GMMs have been employed to model the input observations, and Bayesian classification has been incorporated to estimate a binary classification of the given GMMs\cite{jasim2014contact}\cite{jasim2017contact}. 
The \emph{expectation maximization}~(EM) algorithm has been demonstrated to be efficient in optimizing the parameters of the given GMMs. 
Jasim et al. utilized the distribution similarity measure~(DSM) to determine the optimal number of GMM components based on the previous work\cite{jasim2017contact}, and this process significantly enhances the modeling performance and computational cost for contact state modeling of the flexible objects. 

HMMs show advantages in recognizing both the contact states and state transitions over the previous contact state classification approaches\cite{hovland1998hmm}\cite{lau2003hmm}\cite{hannaford1991hmm}. 
In this way, the contact state classification problems solved by HMMs are capable of taking the temporal information into account. 
A multiple contact model method incorporated into an HMM model to estimate the contact sequence was proposed in \cite{debus2004hmm}, and this model only requires the partial observations, such as kinematic data without other object information. 
Basically, the aforementioned contact state recognition approaches are supervised learning problems, which require considerable labeled samples and extensive training first. 

In contrast to parametric learning techniques, the \emph{random forest}\cite{cabras2016random} technique,  without determining the parameters in advance,  was explored for multiple classifications. 
In addition, the binary \emph{stochastic gradient boosting}(SGB)\cite{cabras2010contact} classifier,  based on its strength of classifier diversity, can perform the contact state recognition. 
Jasim et al.\cite{jasim2017contact} has compared the success rate and computational time of several most frequently used classification techniques through an assembly experiment with the flexible rubber manipulated objects. 
Based on the given results, 
we provide a comprehensive summary shown in Table.~\ref{table_staistical_model}, covering the comparison of the success rate, computational time and pros/cons of the introduced statistical techniques for contact state recognition. 

In conclusion, for real-world robotic peg-in-hole assembly with a limited number of samples, NNs and FC-based methods cannot handle complicated contact model recognition well. 
Nevertheless, GMMs and SVM have shown the high efficiency and better generalization for contact state classification. 
Furthermore, HMMs can cooperate with other classification algorithms to take the effect of state transition into account. 


\subsection{Compliant control}

In contrast to the general assembly\cite{lefebvre2005active}\cite{2012contactsvm}, the contact model-based control strategies for peg-in-hole assembly depicted in \reffig{fig_control_framework} can be simplified into two steps: a high-level planning module and a low-level controller. 
The high-level planning module is used to derive the high-level commands for low-level controllers based on based on the geometrical and environmental constraints. 
The low-level controller with the set of high-level commands is used to execute the assembly actions according to the observed wrench signals and pose information and the current contact state. 

\subsubsection{Low-level controller}
At present, the robots are able to handle the point-to-point accuracy requirements easily, and the position controller has become quite mature\cite{lopes2008force}. 
The low-level controllers of industrial robots generally consist of two categories: force-based impedance controllers and position-based force controllers. 
The force-based impedance controllers aim to execute the commands for joint torque\cite{ren2018learning}, and the measured wrench signals are used to generate the desired torque value for the inner force loop. 
The position-based force controllers typically generate the desired position and orientation commands according to the outer force loop; then, the commands are executed by the inner position controller. 
Both of these controllers have strengths and weaknesses; nevertheless, direct access to actuator torques data is not available for most industrial robots. 
The position-based force controllers are widely utilized for the industrial assembly control through the designed outer force controller\cite{kuangenjamming}\cite{hou2018learning}\cite{tang2016autonomous}. 

To accommodate the environmental uncertainties of the assembly process, some researchers have focused on optimizing the parameters of the position-based force controller. 
A network-based adaptive fuzzy model guided by the contact state estimator has been proposed to acquire adaptive parameters for a force impedance controller\cite{1998conceptfuzzy}. 
To optimize the outer proportional-integral-derivative~(PID) force controller through few trials for real-world peg-in-hole assembly, Hou et al. \cite{hou2018learning} proposed \emph{evolutionary algorithms}~(EA) in conjunction with \emph{support vector regression}~(SVR) to obtain the optimal PID parameters. 

\subsubsection{High-level planning module}

For peg-in-hole assembly, the high-level planning module generally generates the desired force and moment value for the low-level force controller according to geometrical constraints of mating parts\cite{zhang2017force}\cite{kuangenjamming}\cite{hou2018learning}. 
In addition to the geometrical constraints, 
Qiao et al.\cite{qiao2015concept} took environmental constraints into account based on the concept of ARIE, as shown in \reffig{fig_control_framework}, and the position uncertainties were eliminated by coarse wrench signals. 
In \cite{qiao2017iros} and \cite{qiao2015arie}, the constraint region in configuration space and physical space has been discussed, and a two-step insertion strategy and a human-inspired compliant strategy based on the ARIE concept have been verified in a broader range of peg-in-hole tasks with sensor-less systems. 
The environmental constraints based on ARIE can not only compensate for the limitations of the force sensors for high-precision assembly, but also provide the guarantees of safety and reliability in real-world assembly. 
The generality and robustness of the compliant control system have been improved with the assistance of environmental constraints. 

Instead of optimizing the low-level controller directly, some researchers have made significant efforts to optimize the high-level planning module according to the contact model recognition results. 
Son \cite{son2001neuralfuzzy} utilized fuzzy set theory to manage and address the uncertainties according to the prior assembly knowledge. 
Additionally, a neural network was constructed to approximate the nonlinear relationship between the jamming analysis and the insertion control strategy. 
Xia et al.\cite{xia2006dynamicanalysis} proposed a no-jamming and no-wedging assembly strategy by choosing the appropriate set of applied forces and moments based on the corresponding control law for different contact states. 
In \cite{shirinzadeh2011hybrid}, a hybrid methodology was proposed by choosing the corresponding low-level controller according to distinguishing the different contact states. 
Tang et al.\cite{tang2016autonomous} proposed an autonomous alignment method to correct the initial pose before the \emph{inserting} phase based on the estimated contact state. 
As assembly strategies have become more advanced, 
Zhang et al.\cite{kuangenjamming} established jamming diagrams based on contact state analysis for a complicated flexible dual peg-in-hole assembly. Then, a jamming theory was applied to establish the parameters of the low-level PD force controller. 

However, the high-level planning module of peg-in-hole assembly is sometimes neglected, and most studies commonly focus on contact model recognition and low-level control strategies\cite{zhang2017force}. 
Furthermore, it remains unclear how best to optimize the high-level planning module and the low-level controller according to the contact model recognition. 
A flexible and adaptable assembly strategy should match the real-time uncertainties via a smooth integration of all the separate modules. 


\section{Contact model-free learning strategies}\label{S:4}
In contrast to the contact model-based control strategies dealing with the contact state recognition and compliant control separately, the contact model-free strategies combine these two steps together. As shown in \reffig{fig_architecture_this_paper}, contact model-free strategies consist of two categories: LFD and LFE. 

\subsection{LFD}

\begin{figure}[!t]
\centering
\subfigure[]{\includegraphics[width=3.2in]{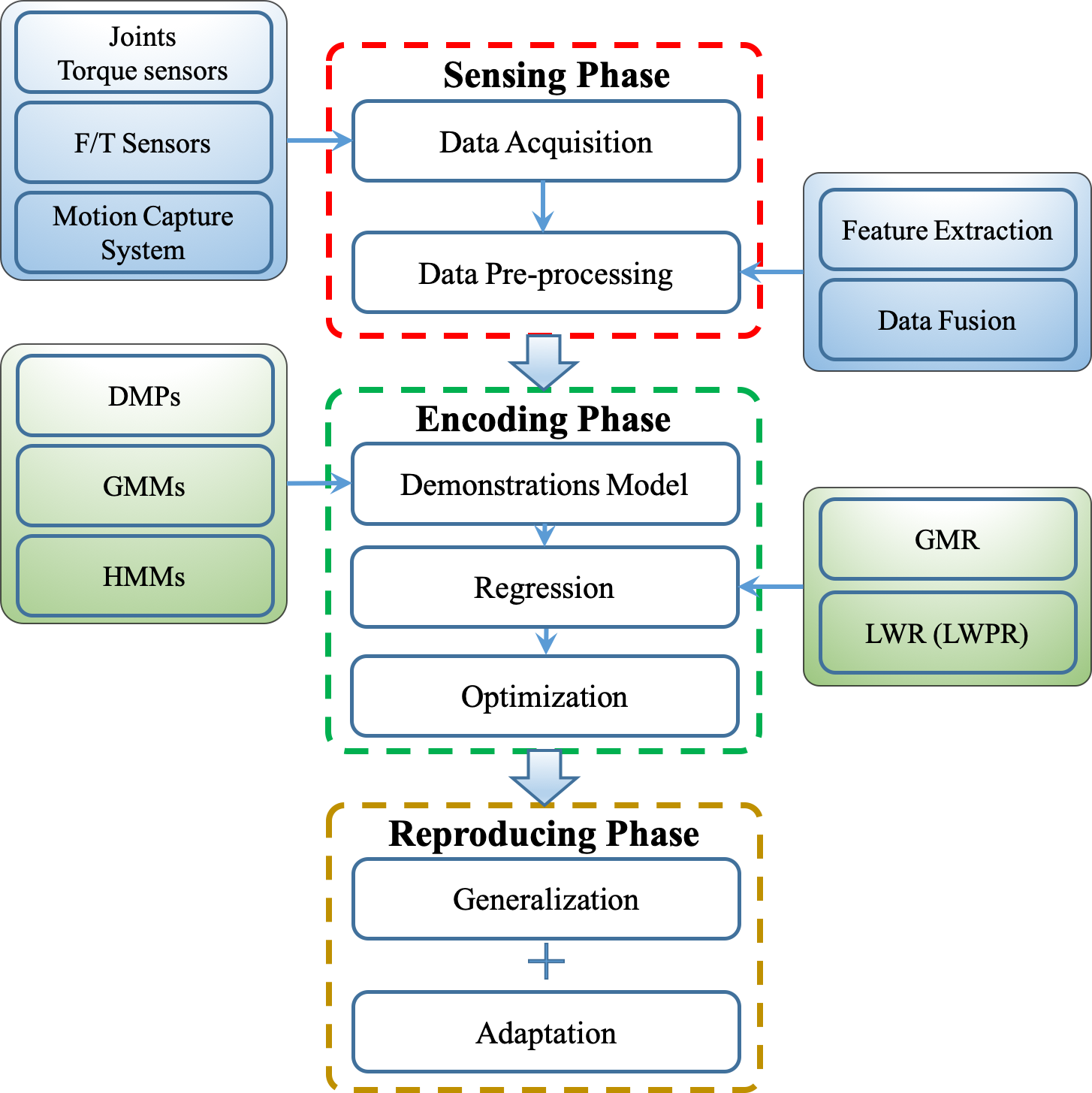}%
\label{fig_framework_lfd}}
\hfil
\subfigure[]{\includegraphics[width=3.2in]{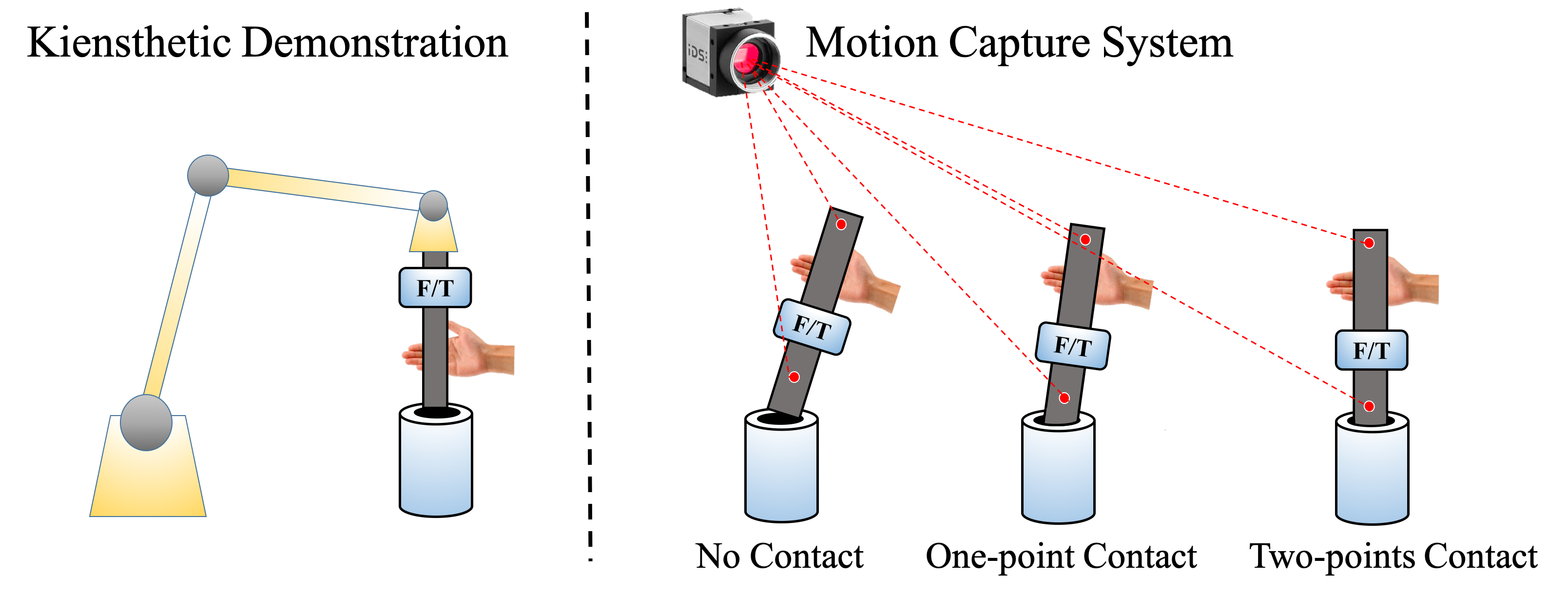}
\label{fig_assembly_demonstrations}}
\caption{Learning from demonstrations strategies. (a)~Framework of the LFD system. (b)~Demonstration approaches.}
\label{fig_learning_from_demonstrations}
\end{figure}

Compared to the industrial robots, humans can perform peg-in-hole assembly with any degree of pose uncertainty due to the flexibility of the wrists, the sensing system and intelligent decision-making ability. 
Instead of analyzing how do humans accomplish the assembly tasks, many researchers focus on simulating the human assembly demonstrations directly and then transforming the skills into robots programming, which is referred as LFD~(also termed \emph{imitation learning} and \emph{apprentice learning}). 
For robotic peg-in-hole assembly, LFD methods consist of three principal phases: sensing, encoding and reproducing, which are depicted in \reffig{fig_framework_lfd}\cite{zhu2018robot}\cite{kyrarini2018robot}. 

\begin{table*}
\centering
\caption{Comparison for learning from demonstrations approaches.}
\label{table_learning_from_demonstrations}
\begin{tabular}{c l l l}
\hline
\textbf{LFD Approaches} & \textbf{Advantages} & \textbf{Disadvantages} & \textbf{References}\\
\hline
DMP 
& $\bullet$ Robust to spatial perturbation 
& $\bullet$ Delay and pause of motion & \cite{2010DMP}\cite{park2008movement}\cite{paxton2015incremental}
\\
& & $\bullet$ Solve multivariate data separately &
\\
& & $\bullet$ Learn from single demo&
\\
GMM 
& $\bullet$ Model joint probability density function& & \cite{kyrarini2018robot}\cite{tang2015learning} \\
& $\bullet$ Handle different source and missing data& &\\
& $\bullet$ Offline learning and online fast regression& &
\\
HMM 
& $\bullet$ Handle partial demonstrations & $\bullet$ Stability sensitive to gains & \cite{1996hmm}\cite{calinon2010learninghmm}\cite{calinon2010overview}\\
& $\bullet$ Handle temporal variability& & \\
& $\bullet$ Handle periodic and reaching movements together & & \\
& $\bullet$ Encode multivariate motion simultaneously&  & \\
\hline
\end{tabular}
\end{table*}
\subsubsection{Sensing phase}This stage aims to interpret the human motion trajectories, including the observed states and executed actions. 
At present, the human demonstration data can be collected through external sensors, such as  
kinesthetic demonstration, motion capture systems, and teleoperated demonstration, which are shown in \reffig{fig_assembly_demonstrations}. 
In this paper, the states of the assembly process, including pose information and wrench signals, can be recorded through \emph{kinesthetic demonstration} and \emph{motion capture systems},
 as shown in \ref{fig_assembly_demonstrations}. 
The pose of pegs can be calculated by robot joints encoder or determined by vision-based pose estimation approaches, such as the extracted 2D boundary features, maker point matching\cite{wan2017optimal} or 3D point cloud data processing\cite{jasim2017contact}. 
The wrench signals are usually detected through the external F/T sensors mounted on the end-effector or joint torque sensors. 
The corresponding executed actions include transnational-rotational offsets or velocities and applied forces acquired through torque sensors or external F/T sensors. 

Additionally, to enhance the performance of interpretation, some data preprocessing techniques are applied for coping with the raw data, such as \emph{principal component analysis}~(PCA) to reduce dimensionality and \emph{dynamic time warping}~(DTW)\cite{song2016guidance} methods to temporally align all sample points from different demonstrations. 
To integrate different types of sensing information, a data fusion architecture\cite{2004datafusion} based on \emph{artificial neural networks}~(ANNs) is used to combine the pose information and wrench signals, and
\emph{kalman filters}~(KF) is utilized to minimize the effects of noise. 

\subsubsection{Encoding phase}
The encoding phase involves mapping the relations among the observed states and the executed actions. 
The mapping approaches have been developed in three main methodologies: \emph{dynamic movement primitives}~(DMPs), \emph{Gaussian mixture regression}~(GMR) and HMMs. 

DMPs, as a nonlinear dynamic system, are utilized to model the discrete movements of the assembly trajectories with sequence of specific goal positions. 
A second-order differential equation is employed to encode the desired movement primitives of assembly trajectories~(positions, velocities and accelerations)\cite{2010DMP}. 
The one component motion specified in joint or task space of the observed state $\mathbf{S} \in \mathbb{R}^{(D_{\mathbf{S}} \times K \times T)}$ is formulated as follows
\begin{equation}
    \mathcal{D} = \{ s^k(t_j), \dot{s}^k(t_j), \ddot{s}^k(t_j);|k=1,...,K; j=1,...,T_k\}
\end{equation}
where $D_{\mathbf{S}}$ denotes the dimensions of the observed state; $K$ denotes the number of demonstration trajectories; $T$ denotes the length of one single demonstration; $T_k$ denotes the length of the $k$ trajectory; 
$s^k(t_j)$, $\dot{s}^k(t_j)$ and $\ddot{s}^k(t_j)$ are the real-time positions, velocity and acceleration, respectively, of the trajectory $k$ at time step $j$. 
Generally, the discrete movements and periodic movements can be represented as a first-order equation and a second-order differential equation, respectively, and can be rewritten in one manner as follows:
\begin{equation}\label{equation_dynamic_system}
    \tau^2 \ddot{s}+\alpha_s \tau \dot{s} -\alpha_s \beta_s(g - s) = \mathbf{f}
\end{equation}
\begin{equation}\label{equ9}
    \mathbf{f} = \mathbf{X} \mathbf{w}
\end{equation}

For discrete movements, the $\mathbf{X}$ can be derived through
\begin{equation}\label{equation_discrete}
\mathbf{X} 
= \left[\begin{array}{ccc}
            \frac{\psi_1(x_1)}{\sum_{i=1}^N \psi_i(x_1)}x_1 & \cdots & \frac{\psi_N(x_1)}{\sum_{i=1}^N \psi_i(x_1)}x_1\\
            \cdots & \cdots & \cdots\\
            \frac{\psi_1(x_T)}{\sum_{i=1}^N \psi_i(x_T)}x_T & \cdots & \frac{\psi_N(x_T)}{\sum_{i=1}^N \psi_i(x_T)}x_T\\
\end{array}
\right]
\end{equation}
where $\psi_i(x) = \mathrm{exp}(-h_i(x-c_i)^2)$ is a radial-basis function; $x = \mathrm{exp}(-\alpha_x t/\tau)$ is a phase variable to guarantee $x$ tends to 0 as time increases. For the periodic movements, the $\mathbf{X}$ can be derived through
\begin{equation}\label{equation_periodic}
\mathbf{X} 
= r \left[\begin{array}{ccc}
            \frac{\varpi_1(x_1)}{\sum_{i=1}^N \varpi_i(x_1)}x_1 & \cdots & \frac{\varpi_N(x_1)}{\sum_{i=1}^N \varpi_i(x_1)}x_1\\
            \cdots & \cdots & \cdots\\
            \frac{\varpi_1(x_T)}{\sum_{i=1}^N \varpi_i(x_T)}x_T & \cdots & \frac{\varpi_N(x_T)}{\sum_{i=1}^N \varpi_i(x_T)}x_T\\
\end{array}
\right]
\end{equation}
where $\varpi_i(x) = \mathrm{exp}(h_i(\mathrm{cos}(x - c_i) - 1))$; the phase variable $x$ moves with the constant speed $\Omega = 1/\tau$; $r$ is the amplitude of the oscillator;  
$\tau$ denotes the period of periodic movements or duration of the training movement; $g$ denotes the target position. Furthermore, $\mathbf{f}$ denotes a nonlinear function representing the convergence property of the position $s$ towards the target value with the following two formulations, respectively. $\alpha_s$ and $\beta_s$ are constant and set to ensure the convergence of the dynamic system represented by (\ref{equation_dynamic_system}). 

DMPs are defined by the parameters $\mathbf{w}$, $\tau$ and $g$. $g = y^k(t_{T_k})$ can be set directly according to the samples; the duration of training movements can be chosen as $\tau$; and the parameter $\mathbf{w}$ could be calculated from the solution of (\ref{equ9}) in the recursive least square manner. 
For better regression performance, the multiple variables of DMPs are estimated in a separate process synchronized by the phase variable. 
For instance, \emph{locally weighted regression}~(LWR) with lower computational complexity is applied to synthesize the parameter $\mathbf{w}$ and nonparametric \emph{Gaussian process regression}~(GPR) with high accuracy is applied to estimate $g$ and $\tau$. 
DMPs-based approaches have been applied to reach a target or follow a periodic path by a set of mass-spring-damper mechanisms. 
In \cite{abu2014solving} and \cite{kramberger2017generalization}, a complete methodology is proposed to learn from the human assembly demonstrations by combining DMPs to capture the trajectories of pegs with the force-torque profiles. 
Furthermore, the differential equation of DMPs has been improved to adapt to the uncertainty in the desired position and obstacle avoidance\cite{park2008movement}. 

GMR is introduced to estimate the relation between the observed states and the control commands. 
GMR is a real-time regression solution that it can reproduce the trajectories modeled by a GMM or modified GMM, and the reproduced trajectories can be adapted to control robot assembly tasks. 
In \cite{tang2015learning} and \cite{tang2016teach}, GMR is employed to predict the velocities in a manner similar to the human in response to wrench signals; then, the output velocities are executed through a low-level controller~(impedance controller) to realize the peg-in-hole insertion phase. 
To construct a heavy-weight component assembly process, Wan et al.\cite{wan2017optimal} proposed a complete methodology through learning assembly skills from human demonstrations and compensating for the large deformation with GPR. 
The joint probability distribution $p(\mathbf{A}, \mathbf{S})$ is calculated with a mixture of $N$ Gaussian components weighted by $\alpha^i$ as follows
\begin{equation}
    p(\mathbf{A}, \mathbf{S}) = \sum_{i=1}^N \alpha^i p^i(\mathbf{A}, \mathbf{S})
            = \sum_{i=1}^N \alpha^i \mathcal N(\bm{\mu}^i, \bm{\Sigma}^i)
\end{equation}
where $\mathbf{S}$ is the observed state as discussed above; $\mathbf{A} \in \mathbb{R}^{(D_{\mathbf{A}} \times K \times T)}$ denotes the assembly actions; and $D_{\mathbf{A}}$ denotes the dimensions of the assembly actions. 
Each component $p^i(\mathbf{A}, \mathbf{S})$ features a Gaussian distribution with a mean of $\bm{\mu}^i=[\bm{\mu}_{\mathbf{A}}^i, \bm{\mu}_{\mathbf{S}}^i]^{T}$ and covariance
\begin{equation}
\bm{\Sigma}^i=
\left[ \begin{array}{cc}
\bm{\Sigma}_{\mathbf{AA}}^i & \bm{\Sigma}_{\mathbf{AS}}^i\\
\bm{\Sigma}_{\mathbf{SA}}^i & \bm{\Sigma}_{\mathbf{SS}}^i\\
\end{array}
\right ]
\end{equation}

The conditional probability $p(\mathbf{A}|\mathbf{S})$ is derived by a weighted summation of each  $p^i(\mathbf{A}|\mathbf{S}) \sim \mathcal{N}(\bm{\mu}_{\mathbf{A}|\mathbf{S}}^i, \bm{\Sigma}_{\mathbf{A}|\mathbf{S}}^i)$ as follows:
\begin{equation}
    p(\mathbf{A}|\mathbf{S}) = \sum_{i=1}^N \frac{\alpha^i p^i(\mathbf{S})}{\sum_{j=1}^N \alpha^j p^j(\mathbf{S})}p^i(\mathbf{A}|\mathbf{S})
\end{equation}
where $p^i(\mathbf{S}) \sim \mathcal{N}(\bm{\mu}_{\mathbf{S}}^i, \bm{\Sigma}_{\mathbf{S}}^i)$ is the marginal probability of input variable $\mathbf{S}$. 
Therefore, the parameters$(\bm{\mu}^i, \bm{\Sigma}^i, \alpha^i)$ of $p^i(\mathbf{A}|\mathbf{S})$ can be estimated iteratively based on the collected demonstration training data by calculating maximum likelihood estimation through the EM algorithm. Then, with the learned Gaussian parameters, the optimal predicted output $\hat{\mathbf{A}}$ could be calculated by maximizing $p(\mathbf{A}|\mathbf{S})$ as follows
\begin{equation}\label{equ4}
    \hat{\mathbf{A}} = \sum_i^N h_i(\mathbf{S}) (\bm{\mu}_{\mathbf{A}}^i + \Sigma_{\mathbf{AS}}^i(\Sigma_{\mathbf{S}}^i)^{-1}(\mathbf{S} - \bm{\mu}_{\mathbf{S}}^i))
\end{equation}
where the weights $h_i(\mathbf{S})$ could be calculated by 
\begin{equation}\label{equation_weights}
    h_i(\mathbf{S}) = \frac{\alpha^i \mathcal N(\bm{\mu}_{\mathbf{S}}^i, \bm{\Sigma}_{\mathbf{S}}^i)}{\sum_{j=1}^N \alpha^j \mathcal N(\bm{\mu}_{\mathbf{S}}^j, \bm{\Sigma}_{\mathbf{S}}^j)}
\end{equation}

Therefore, the GMR can be learned offline and the learned regression function calculate the expected actions $\hat{\mathbf{A}}$ rapidly online, which makes it appropriate to perform the assembly in real-time. 

HMMs have been extensively used to encode and generalize the observed assembly trajectories of humans due to the strengths of the spatial and temporal variability\cite{1996hmm}\cite{calinon2010learninghmm}\cite{calinon2010overview}. 
HMMs considered as a type of dynamic Bayesian network and are employed to model the real state transition in assembly processes~(vision feedback and force-moments). 
An HMM model generally includes five components, hidden state $\mathbf{S}$, observable state $\mathbf{O}$, initial state probability matrix $\bm{\Pi}$, hidden state transition probability $\bm{\Psi}$ and observable state transition probability matrix $\bm{\Omega}$. 
$\Psi_{ij}= p(\mathbf{S}_j|\mathbf{S}_i)$ represents the probability of the state transition from $i$ to $j$, and $\Omega_{ij} = p(\mathbf{O}_j|\mathbf{S}_i)$ represents the probability of acquiring the observation $\mathbf{O}_j$ at the state $\mathbf{S}_j$~(always $1$ for real-world peg-in-hole assembly). 
The joint probability $p(\mathbf{A}, \mathbf{S})$ is encoded by HMM model with a continuous $K$ state and each state is encoded by GMR with mean $\bm{\mu}$ and convariance $\bm{\Sigma}$. 
Therefore, the HMM model can be defined by parameters $(\bm{\Pi}, \bm{\Psi}, \bm{\mu}, \bm{\Sigma})$, which can be learned by the EM algorithm\cite{1996hmm}. 
Compared to the original GMR approach in (\ref{equ4}), the weight $h_i(\mathbf{S})$ representing the importance of different Gaussian is constant, which is extended to in \cite{calinon2010learninghmm} by recursively calculating a maximum likelihood represented as the HMM model. 
The weight $h_i(\mathbf{S})$ can be derived as 
\begin{equation}
    h_{i,t}(\mathbf{S}) = \frac{\alpha^i (\sum_{k=1}^K h_{k, t-1}(\mathbf{S}) \Psi_{ki}) \mathcal N(\bm{\mu}_{\mathbf{S}}^i, \bm{\Sigma}_{\mathbf{S}}^i)}{\sum_{j=1}^N \alpha^j (\sum_{k=1}^K h_{k, t-1}(\mathbf{S}) \Psi_{ki}) \mathcal N(\bm{\mu}_{\mathbf{S}}^j, \bm{\Sigma}_{\mathbf{S}}^j)}
\end{equation}
which takes the temporal influence of the dynamic assembly movements into account. 

In conclusion, the comparison of different LFD encoding methods for robotic assembly has been investigated in \cite{zhu2018robot}, \cite{kyrarini2018robot}, \cite{calinon2010overview} and \cite{calinon2010learninghmm}, and our conclusions regarding the pros and cons of these three LFD encoding methods for peg-in-hole assembly are as shown in Table.~\ref{table_learning_from_demonstrations}. 
A significant strength of DWPs is their adaptability to the perturbations through a second-order system. 
GMMs can model the mapping function well with clustering and probability density estimation with high robustness to environmental noise. 
HMMs, encapsulating the precedence information with a state transition metric, can perform imitation learning with partial demonstrations\cite{calinon2010learninghmm}. 
Additionally, in contrast to the DMPs, which require two different equations for periodic and discrete problems, HMMs exploit a unified formulation. 
To enhance the adaptation of learned assembly strategies, many researchers have investigated the
variants of the above modeling methods or have combined them. 
Modified GMMs combined with optimal control algorithms were proposed in\cite{kyrarini2018robot}. 
GMMs combined with HMMs have been explored and have shown competitive performance for robotic assembly\cite{calinon2010learninghmm}. 

\subsubsection{Reproducing phase}

After demonstrations are encoded and regression functions are optimized, the desired assembly actions are reproduced in the reproducing phase. 
The generalization of the learned assembly skills depends on the regression performance. 
Instead of generalizing the motions with statistical regression methods, such as LWR and \emph{locally weighted projection regression}~(LWPR), directly, 
GMR derives the regression function with the joint probability density of collected demonstration data. 
However, the existing LFD methods are at the trajectory level, which is difficult to apply to reproduce more complicated assembly tasks with larger uncertainties. 
Additionally, the generalization of new circumstances and the robustness against perturbation in addition to reproducing actions require further improvement. 


\subsection{LFE}

\begin{figure}[!t]
\centering
\subfigure[]{\includegraphics[width=3.in]{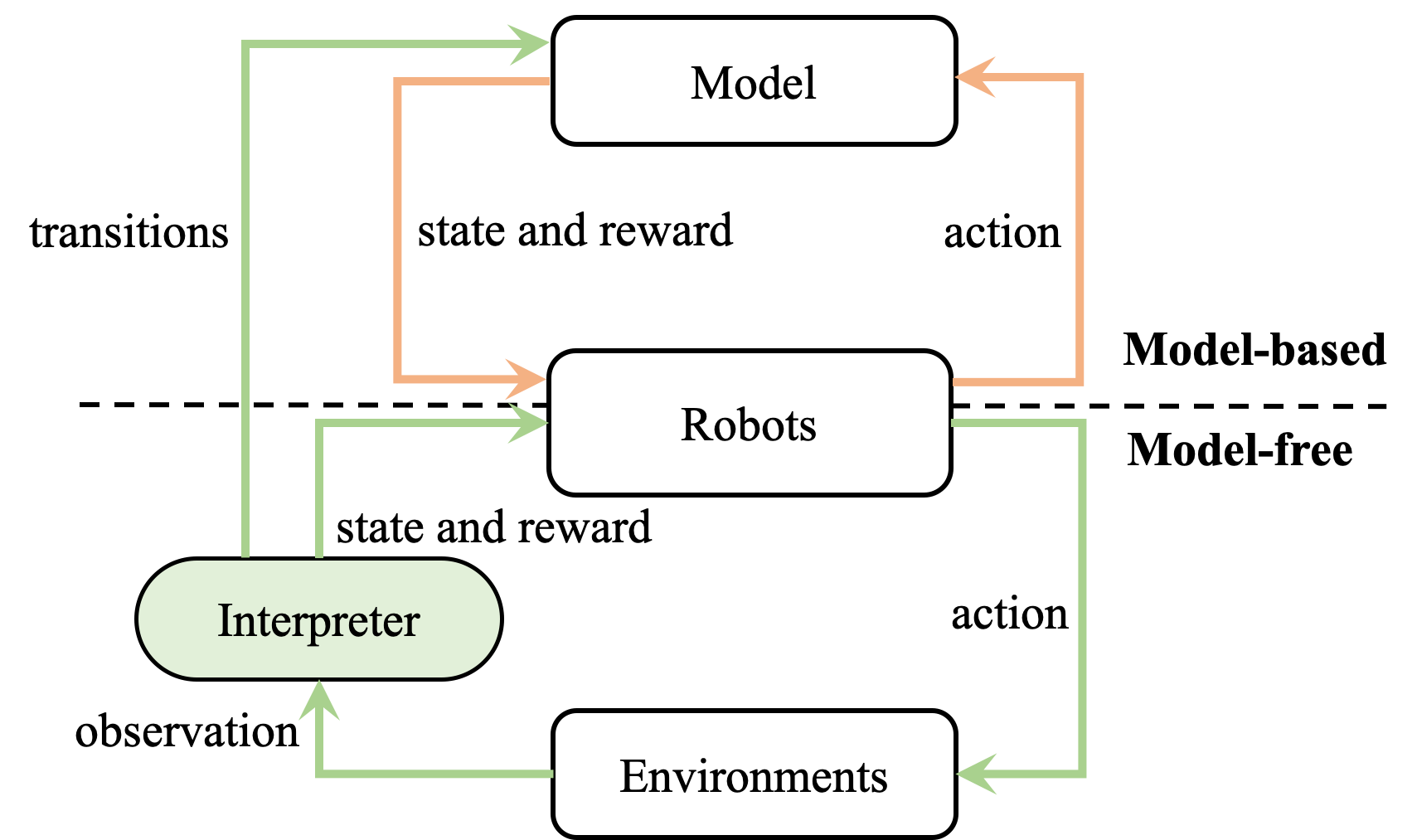}
\label{fig_framework_rl}}
\subfigure[]{\includegraphics[width=3.5in]{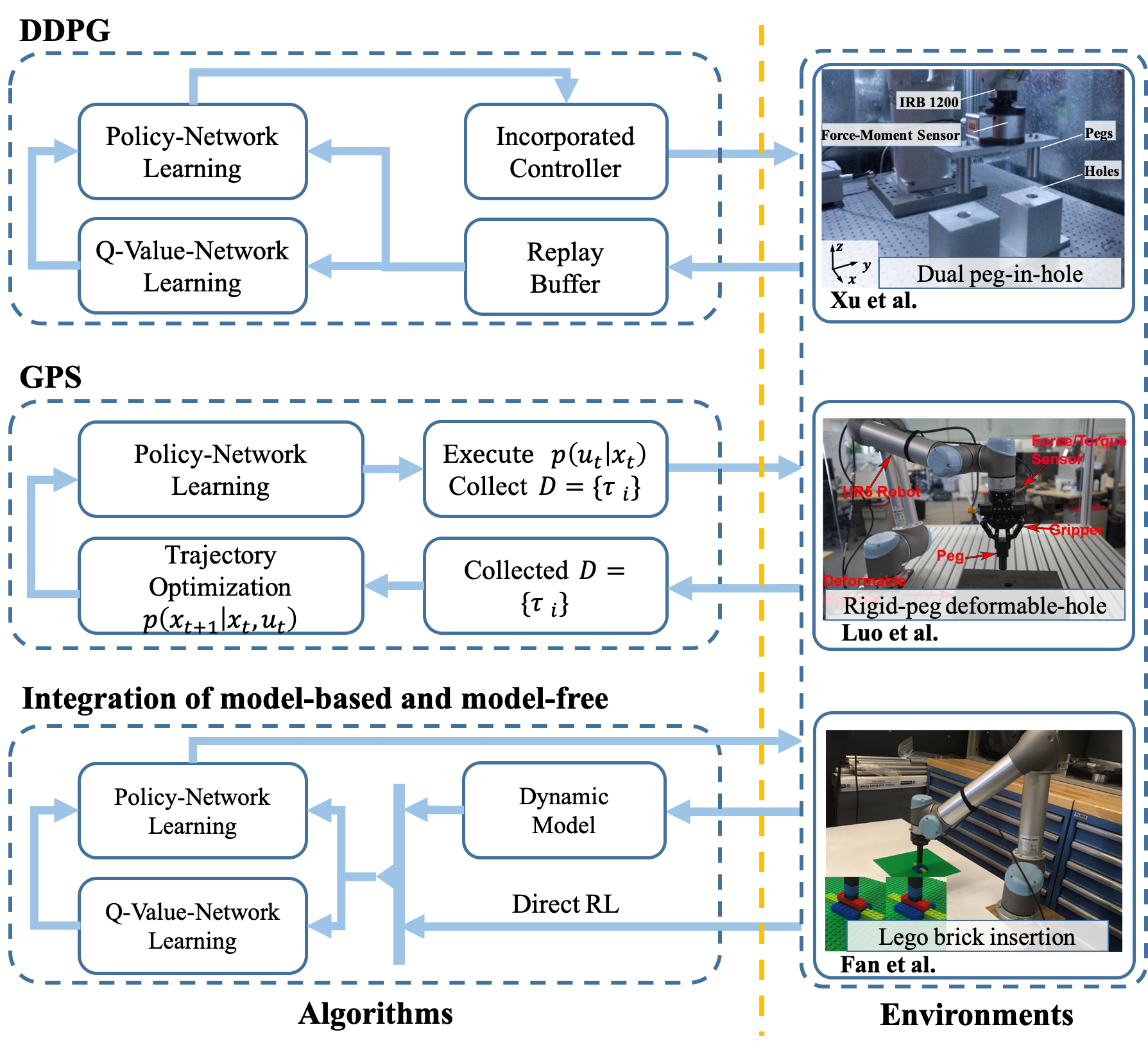}
\label{fig_examples_rl}}
\DeclareGraphicsExtensions.
\caption{(a)~Architecture of reinforcement learning algorithms. (b)~Examples of reinforcement learning applied in robotic peg-in-hole assembly.(Xu et al.\cite{zhiminfeedback}, Luo et al.\cite{rl2018peginhole} and Fan et al.\cite{2018modelfreelearning}} 
\label{fig_learing_from_environments}
\end{figure}

The development of highly intelligent control systems with the ability to learn skills autonomously has advanced considerably. 
A promising direction based on RL has been extensively used to solve challenges related to complicated contact-rich assembly tasks\cite{rlhighpercision}\cite{zhiminfeedback} \cite{rl2018peginhole}. 
The core idea of RL-based strategies is that the robot learns and explores the assembly policy actively given a high-level specification of what to do through the reward interpret mechanism instead of guiding the specific actions explicitly. 
Furthermore, the robots can achieve \emph{incremental learning} by interacting with the environment through the smooth combination of the contact model recognition and compliant control process. 
Recent advances in RL have achieved great success in solving robotic manipulations issues, especially in conjunction with \emph{deep neural networks}～(DNNs) for parameterizing policies and value functions. 
As shown in \reffig{fig_framework_rl}, RL approaches are generally distinguished into typical model-free and model-based two main classes according to whether there is a learned model of the dynamic transitions between the robot and environment. 
Additionally, the integration of model-based and model-free techniques has also drawn considerable attention in recent years. 

\begin{table*}[!t]
\centering
\caption{Comparison for reinforcement learning algorithms.} 
\begin{tabular}{l l l l}
\hline
\textbf{Category} & \textbf{Advantages} & \textbf{Disadvantages} & \textbf{Methods}\\
\hline
Model-free & $\bullet$ No need for prior knowledge of environment & $\bullet$ Performance depends on transition model & DDPG\\
           & $\bullet$ Easy implementation & $\bullet$ Divergence due to the bias of model & DQN~(Q-learning)\\
Model-based & $\bullet$ Fewer interactions with environments & $\bullet$ Less data efficiency & GPS\\
            & $\bullet$ Fast convergences to optimal policy & $\bullet$ Unstable and dangerous & PILCO\\
\hline
\end{tabular}
\label{table_comparison_rl}
\end{table*}

\subsubsection{Model-free RL} 
Model-free RL methods aim to learn the optimal policy by simultaneously exploring the state-action space and estimating a dynamic model from the transitions simultaneously\cite{jan2013reinforcement}. 
The solution for solving an RL problem can be decomposed into two alternative method families: \emph{value-based} methods and \emph{policy-based} methods. 
A value-based method was proposed for the first time to learn the value function through the nonlinear function approximation~(via \emph{NNs}), and the discrete assembly action was chosen in a $\epsilon-$greedy manner\cite{gullapalli1992learning}. 
A value-based learning algorithm with a \emph{long short-term memory}~(LSTM), a variant of a \emph{recurrent neural network}~(RNN) to estimate the $Q$ value function in order to achieve peg-in-hole assembly with a precision exceeding the resolution of the robots\cite{rlhighpercision}. 
Additionally, a learning framework for solving the real-world robotic assembly problem was proposed: the output of the RL control system was used as the settings of the low-level position-based force controller instead of controlling the robots directly. 

The limitation of the value-based methods is that the output actions of the RL system can only be discrete and low-dimensional. 
Policy-based methods have been extensively explored in the case of high-compliance robotic applications. 
The RL method implemented with actor and critic two components was proposed to derive the assembly policy for the actor and critic and was used to evaluate the actions\cite{nuttin1997learning}. 
As policy-based methods have advanced, \emph{deterministic policy gradient}~(DPG) theory was derived in \cite{silver2014deterministic} to achieve differential policy learning with a high stability. 
Subsequently, \emph{deep deterministic policy gradient}~(DDPG) approaches\cite{duan2016benchmarking} have been developed through combination with DNNs, and these approaches have been widely applied for high-compliance continuous action control applications\cite{zhiminfeedback}\cite{ren2018learning}. 

Policy-based methods are appropriate to solve the real-world problems with the continuous and high-dimensional actions. 
The learning of the parameterized policy always converges slowly with a high degree of variance and instability. 
To date, some studies on improving the stability and efficiency of the DDPG framework for real-world robotic assembly have been published, which allows learning from different samples distribution in an off-policy scheme. 
A model-driven DDPG algorithm was proposed to learn the general assembly policy for multiple peg-in-hole problems\cite{zhiminfeedback}. As shown in \reffig{fig_examples_rl}, one contribution of the model-driven DDPG algorithm is that the learning of the actor network is driven by the basic actions from the simple but practical controller. 
Additionally, many research studies have focused on incorporating prior knowledge to enhance the efficiency. 
A DDPG from demonstration~(DDPGfD) method was proposed in \cite{vecerik2017leveraging} by inputting human demonstrations into the expert memory buffer, which are reused by a prioritized replay mechanism to enhance policy learning. 
In contrast to providing a baseline policy with for robots, prior knowledge about the geometric information of assembly parts was used to plan the motion trajectory in \cite{2018learningfromCAD} to guide the policy learning. 
Basically, those authors focused on the assembly motion planning with geometrical information from a computer aided design~(CAD) and utilized the RL algorithm to handle the dynamics of the environment. 

\subsubsection{Model-based RL}

In contrast to the typical model-free RL methods, model-based methods aim to learn a dynamic model with the stored transitions, and the policies are optimized by deriving the rewards and next state from the learned model\cite{polydoros2017survey}. 
For complicated manipulations tasks, \emph{policy search} methods for deriving the optimal policies through interacting with the learned dynamic model directly have shown faster convergence. 
\emph{Guide policy search}~(GPS) has been developed to learn a couple of manipulation behaviors, as shown in \reffig{fig_examples_rl}; this method combines a trajectory optimization component and a neural network policy learning component\cite{levine2015learning}. 
Luo et al. proposed mirror descent GPS~(MDGPS) to tackle a complicated assembly task with rigid pegs and deformable holes for use with noncompliant robots and external F/T sensors\cite{rl2018peginhole}. 
The \emph{probabilistic Inference for Learning Control}~(PILCO)\cite{deisenroth2011pilco} framework employs a \emph{Gaussian Process} to model the transition dynamics and a linear function to represent the policy, and this framework is a state-of-the-art model-based RL algorithm in terms of the sample efficiency and time efficiency. 

Consequently, model-based RL methods only need to explore a narrower space than the model-free methods, resulting in faster convergence with fewer interactions with the environment. 
However, the performance of model-based RL methods heavily depends on the accuracy of the learned transition dynamic model. 
Polydoros and Nalpantidis gave an up-to-date overview of model-based RL algorithms and the related robotic applications in \cite{polydoros2017survey}. 
We summarize the pros and cons of the model-free and model-based RL algorithms for robotic peg-in-hole assembly as shown in Table.~\ref{table_comparison_rl}. 

\subsubsection{Integration of model-based and model-free methods}

Both model-free and model-based RL methods have advantages and disadvantages, as summarized in \cite{polydoros2017survey}. 
Model-free RL methods can perform the complicated assembly problems prominently with a general and easy implementation way but are less efficient. 
DDPG-based model-free algorithms can provide more stable policies and attain the asymptotic performance in some assembly tasks that exceeds the performance of nonsmooth dynamics models. 
Model-based RL methods are able to enhance policy learning by utilizing rich transition information. 
Additionally, model-based optimal controllers constrain the exploration space to a safe region but often cannot consistently achieve good convergence performance due to a large model bias. 
In \cite{2018modelfreelearning}, Fan et al. analyzed a model-based RL method~(GPS) and a model-free RL method~(DDPG), and then proposed a more efficient framework by combining the model-based optimal control strategies with a model-free actor-critic based learning algorithms, as shown in \reffig{fig_examples_rl}. 

Recently, the integration of the strengths of model-based and model-free RL methods has been a well-studied topic for decades. 
Most of the efforts have focused on smoothing the transition from model learning to policy learning and obtaining more useful information from sample transitions. 
In \cite{pong2018temporal}, the authors introduced a novel strategy called as the \emph{temporal difference} model~(TDM) by training a goal-conditional value function with a specific choice of reward and horizon prediction. This model made the robots consider not only reaching a goal state as optimally as possible but also as easily as possible. 
The TDM conditions was extended in a multistep model study\cite{venkatraman2016multimodel} to not only predict a sequence state in the future but also to reach a possible goal state based on the \emph{general value functions} idea in\cite{sutton2011horde} by learning rich contextual value functions from one single experience dataset. 
Additionally, some researchers have focused on exploring how to make full use of the learned dynamic model in addition to the commonly used Dyna architecture\cite{sutton1991dyna} and GPS-based methods\cite{levine2015learning} in order to simulate the entire trajectory every iteration. 

\section{Discussion and conclusion}\label{S:5}

We have surveyed the remarkable work on robotic peg-in-hole assembly processes and have provided a comparison of different strategies summarized in Table.~\ref{table_final_comparison}. 
Both contact model-based and contact model-free strategies can achieve distinguished performance in some special scenarios. 
In summary, contact model-based conventional controllers and LFD methods
guarantee safety and efficiency and are suitable for special assembly scenarios after adjustment with preprogramming beforehand. 
LFE algorithms based on RL are promising for actively and flexibly performing a broad range of complicated assembly process. 
Similar to human beings decision-making systems without tedious programming and rules, RL-based algorithms can remove the specificity engineering of the feedback controller, and they can naturally solve assembly problems with large environmental uncertainties and generalize to new situations.  

\begin{table}[!t]
\centering
\caption{Conclusion for Robotic peg-in-hole assembly strategies.}
\begin{tabular}{l c c c}
\hline
\multirow{2}{*}{\textbf{Category}} & \textbf{Contact model-based} & \multicolumn{2}{c}{\textbf{Contact model-free}}\\
\cline{3-4}
                  & & LFD & LFE\\
\hline
Pre-programming & $\checkmark$ & $\times$ & $\times$\\
\cline{1-4}
Data-efficiency & $\checkmark$ & $\times$ & $\times$\\
Safety-guarantee & $\checkmark$ & $\checkmark$ & $\times$ \\
Generalization & $\times$ & $\times$ & $\checkmark$ \\
\hline
\end{tabular}
\label{table_final_comparison}
\end{table}

It is clear that it is not possible for robots to perform the peg-in-hole assembly as flexibly as human beings based solely on any single strategy.  
Although RL-based contact model-free algorithms have attracted more attention than contact model-based algorithms and LFD methods, 
RL is not the main component for deriving an assembly strategy with sufficient robustness and flexibility to perform all the robotic peg-in-hole assembly problems. 
Furthermore, typical model-free RL-based methods are still not the suitable way for robotic problems. 
Consequently, we highlight a couple of open questions in the field of robotic peg-in-hole assembly and propose some potential directions for future research. 

\subsection{Open questions in the field of robotic peg-in-hole assembly?}

\subsubsection{How can the active compliant control strategies cooperating with passive compliant mechanisms be improved?} 

With the development of sensing hardware and robotic perception techniques, active compliant control strategies have been extensively explored for robotic peg-in-hole assembly. 
In addition, high-compliance robots have also been employed develop complicated assembly systems with a simple compliant strategy. 
Both the improvement of active compliant control strategies and passive compliant mechanisms can promote assembly research. 
For a peg-in-hole assembly, the large position or force uncertainties can be accommodated by an active compliant control strategy, while smaller uncertainties can be eliminated through improving the compliance of mechanism instead of optimizing the parameters of the active controller. 
Therefore, the incorporation of active compliant control strategies and passive devices still requires more attention to decide when to optimize the compliant control strategy or modify the passive mechanism. 

\subsubsection{How can effective and incremental demonstration learning be realized?}

LFD methods provide a solution to perform the robotic peg-in-hole assembly without handcrafted preprogramming according to contact model recognition. 
Although it is challenging to collect demonstration experiences, it is an essential task for improving not only data efficiency but also the adaptation and generalization of the learned assembly policy. 
For instance, in an attempt to solve this challenge, DMPs were used as fundamental blocks with RL to learn advanced skills\cite{rldmp2011reinforcement}. 
RL is commonly used to obtain the adaptive parameters for robust results. 
Additionally, to improve the efficiency, better feature extraction methods are required to select better demonstrations and omit undesirable information. 

\subsubsection{How can model-based and model-free RL algorithms by combined?} 
It is clear that the integration of model-based and model-free RL algorithms is a promising solution to promote the RL based strategies in robotics peg-in-hole assembly, but this issue introduces two key points: how can a perfect dynamic model be learned? and how can robots be made to balance learning from the transition model and learning directly from the environments? 

A good transition model representing the dynamics of the environment allows the robots to have a true understanding of the environment, which ensures that the optimal policy can be chosen accurately based on the model.  
In specific real-world robotic problems, the environment has been explored as a physics-based model and as a statistical model from experience data, including deterministic models and stochastic models. 
In the learning process, the statistical model can be considered as a \emph{supervised learning} problem. \emph{Deep learning} has achieved a major advances in function approximation, but a low sample efficiency still limits the performance in real-world scenarios. 
Therefore, one point is how to consider the environmental uncertainties or the existing physics model in transition model learning. 
Additionally, the transition model can be extended by taking prior domain knowledge, such as expert experience, into account. 

As shown in \reffig{fig_framework_rl}, the robots decide when to interact with the transition model, and the degree of confidence in the transition model greatly affects the quality of the learned assembly skills. 
Therefore, scalable methods for effectively planning based on the given transition model are still required, in addition to the Dyna architecture\cite{sutton1991dyna} and GPS-based methods \cite{levine2015learning}.

\subsection{Potential future work}

To combine the strengths of contact model-based and contact model-free learning algorithms, we propose the following directions to explore the possible solutions in the field of robotic peg-in-hole assembly. 

\subsubsection{Incorporate the knowledge representation method into contact model recognition and transition model learning.} 
Contact model recognition and transition model learning still require better feature extraction methods for the assembly environment. 
A promising solution is a knowledge representation method based on \emph{general value functions}, which was proposed to represent the understanding of environments through learning some simple auxiliary tasks given some prior knowledge. 

\subsubsection{Incorporate prior knowledge into learning process}
Prior knowledge can be the existing control law as in \cite{zhiminfeedback} or can be interpreted through learning from expert demonstrations. 
Additionally, predictions about the environments can be learned as GVFs, which can also be considered as prior knowledge for incorporating into the learning process. 
For instance, prior knowledge can be used to improve the balance of the model-based and model-free RL strategies. 

\subsubsection{Incorporate physical model into reward shaping for RL-based algorithms}

The solution of robotic peg-in-hole assembly problems through RL-based algorithms holds great promise. 
However, most of real-world problems are sometimes difficult to interpret with reward signals, and unpredictable exploration and dangerous actions need to be reduced. 
How well the designed reward mechanism shapes the assembly problems affects the the quality and efficiency of learning. 
For instance, Xu et al.\cite{zhiminfeedback} investigated a fuzzy reward system to take more prior knowledge into account. 
The \emph{Inverse} RL method was utilized to derive the rewards from the observed expert behaviors, thereby exploiting the knowledge of human beings\cite{abbeel2004apprenticeship}. 
Therefore, the physical-model including the geometric information on parts and a mature friction model can be considered as the implicit constraint on the design of reward mechanism. 
Additionally, instead of the constant reward signals, the reward-based mechanism can be updated by evaluating a high-level objective function according to the designer's final goal, which means that the robots receive different evaluation feedback at different stages.






\section*{Acknowledgment}

The authors would like to thank...

\ifCLASSOPTIONcaptionsoff
  \newpage
\fi



\bibliographystyle{IEEEtran}
\bibliography{IEEEexample}
%



%







\end{document}